\documentclass[10pt,twocolumn,letterpaper]{article}

\usepackage[pagenumbers]{cvpr} 
\usepackage{multirow}
\usepackage{xcolor}
\usepackage{bm}
\definecolor{cvprblue}{rgb}{0.21,0.49,0.74}
\usepackage[pagebackref,breaklinks,colorlinks,allcolors=cvprblue]{hyperref}
\usepackage[table]{xcolor}   
\def\paperID{-} 
\def\confName{CVPR}
\def\confYear{2026}

\title{Geo6DPose: Fast Zero-Shot 6D Object Pose Estimation via Geometry-Filtered Feature Matching}

\author{Javier Villena Toro\\
Linköping University\\
{\tt\small javier.villena.toro@liu.se}
\and
Mehdi Tarkian\\
Linköping University\\
{\tt\small mehdi.tarkian@liu.se}
}

\begin{document}
\maketitle
\begin{abstract}
Recent progress in zero-shot 6D object pose estimation has been driven largely by large-scale models and cloud-based inference. However, these approaches often introduce high latency, elevated energy consumption, and deployment risks related to connectivity, cost, and data governance—factors that conflict with the practical constraints of real-world robotics, where compute is limited and on-device inference is frequently required. We introduce Geo6DPose, a lightweight, fully local, and training-free pipeline for zero-shot 6D pose estimation that trades model scale for geometric reliability. Our method combines foundation model visual features with a geometric filtering strategy: Similarity maps are computed between onboarded template DINO descriptors and scene patches, and mutual correspondences are established by projecting scene patch centers to 3D and template descriptors to the object model coordinate system. Final poses are recovered via correspondence-driven RANSAC and ranked using a weighted geometric alignment metric that jointly accounts for reprojection consistency and spatial support, improving robustness to noise, clutter, and partial visibility. Geo6DPose achieves sub-second inference on a single commodity GPU while matching the average recall of significantly larger zero-shot baselines (53.7 AR, 1.08 FPS). It requires no training, fine-tuning, or network access, and remains compatible with evolving foundation backbones, advancing practical, fully local 6D perception for robotic deployment.
\end{abstract}    
\section{Introduction}
\label{sec:intro}

\begin{figure}[t]
  \centering

   \includegraphics[width=\linewidth]{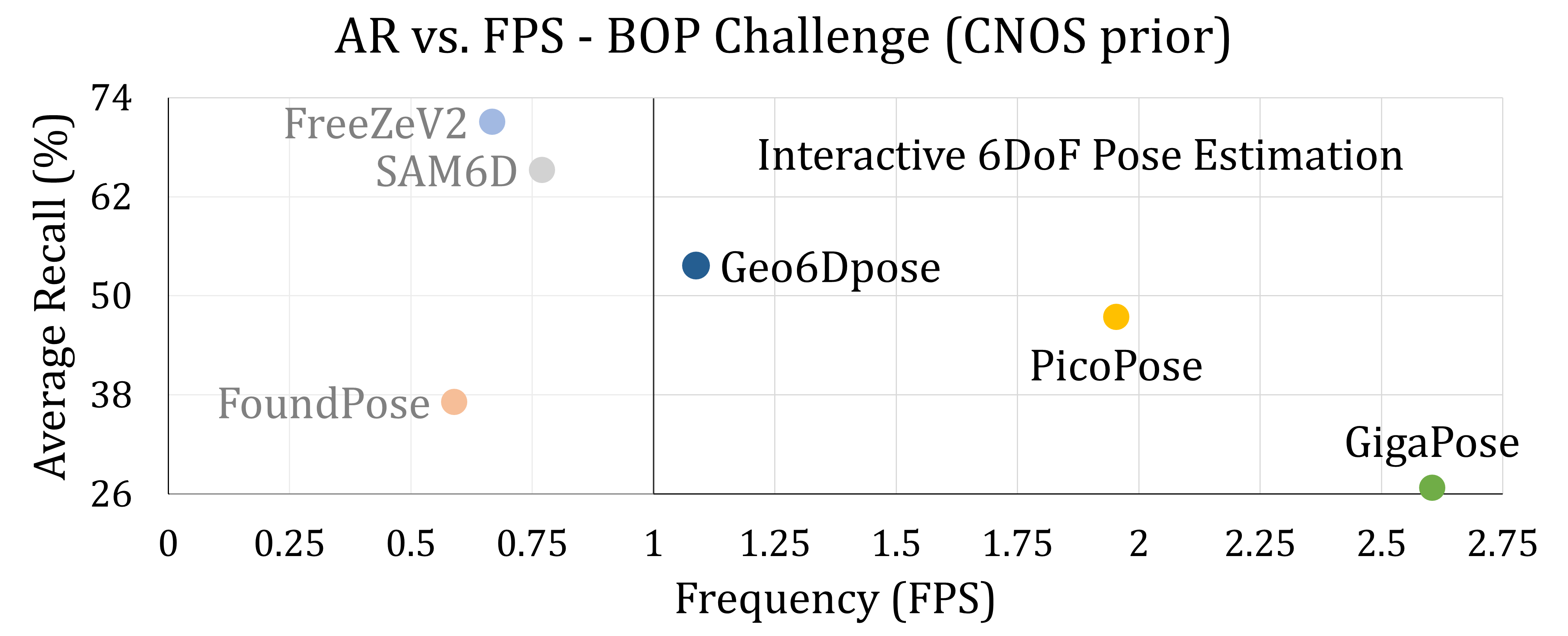}
    \caption{Geo6DPose compared against state-of-the-art zero-shot 6D pose estimation methods using CNOS as the localization prior. The plot shows the trade-off between inference speed (FPS) and Average Recall (AR) averaged over the seven core datasets of the BOP benchmark. Geo6DPose achieves the highest AR among real-time capable methods, operating in the interactive regime while remaining competitive with significantly slower approaches.}
   
   \label{fig:arfps}
\end{figure}

Object pose estimation seeks to recover the 6D transformation—3D position and orientation—of an object relative to the camera or world frame. This capability underpins spatial reasoning across multiple domains: in robotics, it enables precise grasping and manipulation \cite{robo1, robo2}; in AR/VR , it anchors virtual content to real-world geometry \cite{VR1}; in autonomous driving , it supports object tracking and motion prediction \cite{driving1}; and in scene understanding, it provides the geometric context for reconstruction and alignment \cite{recons1}. By linking perception with physical geometry, 6D pose estimation forms a fundamental bridge between visual sensing and action in intelligent systems.

6D pose estimation methods are commonly divided into model-based and model-free approaches. Model-based methods \cite{cosypose, xiang2018posecnn} assume access to a 3D model and estimate the object’s pose by aligning image or depth observations to it, offering precision and interpretability. Model-free methods \cite{he2023onepose} instead learn to regress poses directly from visual input, trading accuracy and generalization for data-driven flexibility \cite{mdpi2024survey}.

Methods also differ by generalization level: instance-level approaches \cite{peng2018pvnet, wang2019densefusion} target specific known objects, category-level methods \cite{SDpose, lin2021dualposenet} generalize within an object class, and zero-shot methods \cite{lin2024sam6d, huang2024matchu, labbé2022megapose} aim to handle entirely novel objects using only their 3D models at inference time. However, most “zero-shot” pipelines still depend on large pretrained or task-specific foundation models (e.g., MegaPose \cite{labbé2022megapose}), leading to heavy compute demands and limited on-device feasibility \cite{nguyen2024gigapose}.

To overcome the data and compute demands of large pretrained or fine-tuned models, recent work has explored training-free pipelines that leverage frozen visual features for correspondence-based alignment. Methods such as FoundPose \cite{ornek2024foundpose} and FreeZe \cite{caraffa2024freeze} use foundation-model descriptors (e.g. DINOv2 \cite{oquab2023dinov2}) extracted from both rendered templates and real images to establish correspondences without any task-specific training. FoundPose, by relying purely on pre-existing visual embeddings enable true zero-shot inference and remain adaptable to new objects or backbones, while FreeZe fuse descriptors from DINOv2 and GeDi\cite{gedi}, a geometric foundation model. Geo6Dpose belongs to this trend of zero-shot train-free methods.

Most modern 6D pose estimation pipelines adopt a coarse-to-fine design: a coarse module first predicts an approximate object pose, typically via global template matching or dense feature correspondence, and a refinement stage then optimizes this pose through local alignment or iterative optimization \cite{nguyen2024gigapose}. While this two-stage structure improves accuracy and robustness, it incurs high computational cost—particularly in the coarse stage, where evaluating thousands of candidate poses or templates dominates runtime. Recent methods such as GigaPose \cite{nguyen2024gigapose} have mitigated this bottleneck by aggressively reducing the search space and employing efficient feature sampling strategies, while PicoPose \cite{liu2025picopose} accelerates inference through a lightweight multi-stage refinement hierarchy. Although both achieve over an order-of-magnitude speed-up compared to prior approaches, these gains come at the expense of reduced pose accuracy.

Geo6DPose addresses this trade-off by introducing a streamlined, training-free pipeline that directly establishes mutual 3D–3D correspondences between scene points and model templates using DINOv2-base descriptors, bypassing costly coarse searches. Our main contributions are:

\begin{enumerate}
    \item Accuracy–Speed Trade-off: Achieving higher average recall than fast trained baselines (e.g., GigaPose \cite{nguyen2024gigapose}, PicoPose \cite{liu2025picopose}) while remaining entirely training-free.
    \item Efficiency: Delivering faster inference than reported training-free methods (e.g., FoundPose \cite{ornek2024foundpose}, FreeZe \cite{caraffa2024freeze}, FreeZeV2 \cite{caraffa2025freezev2}), with competitive accuracy.
    \item Novel Positioning: Occupying a previously unexplored region of the accuracy–runtime Pareto frontier for zero-shot 6D pose estimation.
\end{enumerate}

\section{Related Work}
\label{sec:related}

\noindent\textbf{2D vision Foundation Models -} The shift in 2D visual representation began with models like CLIP \cite{clip}, which leveraged massive image-text pairs to learn a joint embedding space. While pioneering in zero-shot classification, CLIP's representations are optimized for global semantic alignment and are often sub-optimal for dense, fine-grained geometric tasks. This limitation was addressed by Self-Supervised Learning (SSL) methods, notably DINOv2 \cite{oquab2023dinov2}, which demonstrated that a frozen Vision Transformer (ViT) trained on large, curated datasets could produce high-quality, all-purpose visual features that excel at capturing local structure and spatial detail for dense downstream tasks.

Building on this success, DINOv3 \cite{simeoni2025dinov3} (released 2025) scaled SSL to billions of images. The resulting robust, dense, and geometrically-aware features from the DINO family have become a critical component for modern 6D pose estimation, with recent frameworks leveraging DINOv2 priors to adapt 2D features for 3D view-aware representations.

\noindent \textbf{Model-based 6D pose estimation of unseen objects -} The 6D pose estimation landscape is anchored by the render-and-compare paradigm (e.g., MegaPose \cite{labbé2022megapose}), which relies on accurate initial object segmentation. To address this reliance, CNOS \cite{nguyen2023cnos} and MUSE \cite{muse} introduced methods using foundation features (DINOv2, SAM\cite{sam}, FastSam \cite{zhao2023fastsegment}) to generate robust, training-free segmentation priors, improving the crucial localization step.

Making the next conceptual leap, SAM-6D \cite{lin2024sam6d} tightly couples SAM-based segmentation with a two-stage partial-to-partial 3D point matching estimator to produce 6D poses, achieving an excellent balance of speed and accuracy.

In parallel, other specialized methods have emerged: GenFlow \cite{moon2024genflow} focuses on high-accuracy iterative refinement via optical flow prediction; ZeroPose \cite{chen2024zeropose} uses efficient geometric feature matching (DOR pipeline) over rendering for cluttered scenes; FoundationPose \cite{wen2024foundationpose} offers a unified estimation and tracking framework using neural implicit representations; and MatchU \cite{huang2024matchu} improves RGB-D fusion with rotation-invariant descriptors.

Finally, a dedicated axis of work targets coarse-stage efficiency and light-weight inference. GigaPose \cite{nguyen2024gigapose} introduced a fast hybrid template–patch correspondence strategy, while PicoPose \cite{liu2025picopose} continues this trend adopting a three-stage correspondence refinement: a ViT finds the best template, a global affine alignment smooths coarse matches, and a local offset network refines them. These efficiency-focused methods represent the primary competitors to established baselines in modern novel-object 6D pose estimation.

\textbf{Training-free Methods -} All previously mentioned 6D pose estimation methods require some form of training or fine-tuning, either on task-specific data or foundation models. Recently, training-free approaches have emerged that leverage frozen foundation models to directly infer poses without additional learning. ZS6D \cite{ausserlechner2024zs6d} pioneered this zero-shot matching paradigm using self-supervised ViT features. Building on this, FoundPose \cite{ornek2024foundpose} used DINOv2 to establish 2D-3D correspondences via template matching, and DZOP \cite{von2024dzop} replaced ZS6D ViT features with diffusion-based representations from Stable Diffusion.

The FreeZe \cite{caraffa2024freeze} family further advanced training-free pose estimation by combining geometric and visual frozen features, using 3D–3D registration and symmetry-aware refinement. FreeZeV2 \cite{caraffa2025freezev2} improved on this with sparse feature extraction, feature-aware scoring, and a modular segmentation ensemble, achieving faster inference and higher zero-shot accuracy without any training.

\section{Geo6Dpose}
\label{sec:geo6d}

In this section, we describe Geo6DPose, our real-time capable  pose estimation method for unseen objects. A high-level overview of the full pipeline is provided in Section~\ref{sec:over}, followed by detailed descriptions of the core components and key design choices in Sections~\ref{sec:temp}–\ref{sec:pose_hyp}.

\subsection{Problem Definition}
The 6-degree-of-freedom (6DoF) model-based object pose estimation problem consists of recovering the rigid transformation $T = [\mathbf{R} \,|\, \mathbf{t}] \in SE(3)$ of a known 3D object model $\mathcal{M}$ in a scene observation $I$, where $\mathbf{R} \in SO(3)$ and $\mathbf{t} \in \mathbb{R}^3$ denote rotation and translation. The goal is to estimate the pose of all instances of target objects visible in the image. Depending on the assumptions at inference time, the problem is divided into 6D detection, where object localization and pose estimation are solved jointly from the image, and 6D localization, where object instances are assumed to be segmented beforehand.

Geo6DPose operates in the 6D localization setting and targets the unseen object regime, in which test objects are not observed during development. Following BOP Challenge guidelines, unseen-object methods may include an onboarding stage (limited to 5 minutes per object on a single GPU) to build object priors without full training. Additionally, Geo6DPose belongs to a growing class of training-free methods, which rely entirely on foundation models—typically DINO variants—without task-specific learning.

\subsection{Method Overview}
\label{sec:over}

\begin{figure*}[ht]
  \centering
   \includegraphics[width=0.9\linewidth]{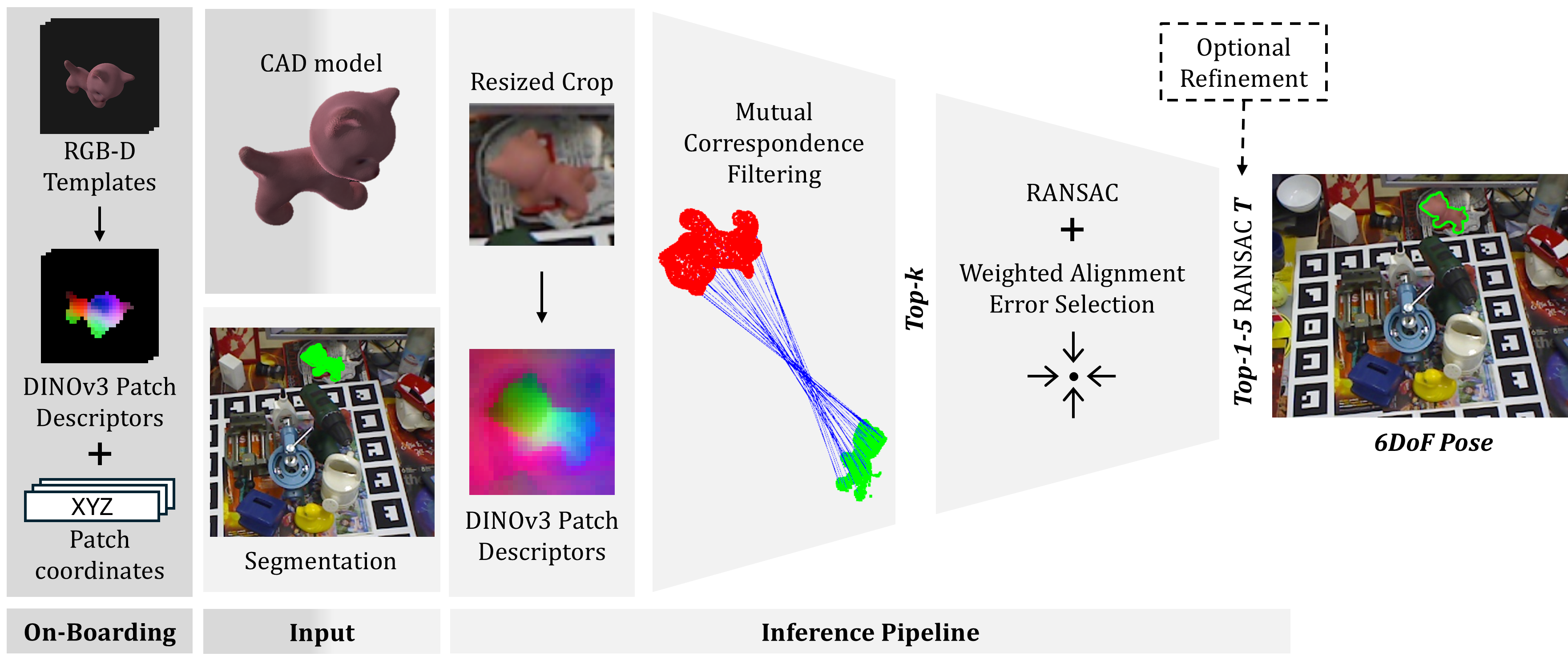}

   \caption{Geo6DPose pipeline. The onboarding stage is executed offline: given a CAD model, pose templates are rendered and DINO \cite{oquab2023dinov2, simeoni2025dinov3} patch descriptors are extracted from RGB \cite{ornek2024foundpose} while patch 3D centers are cached from depth. At inference, a segmentation mask predicted by CNOS \cite{nguyen2023cnos} is used to crop the input frame. Geo6Dpose establishes 3D–3D mutual correspondences by matching template patch coordinates with scene depth based on DINO patch similarity, retaining the top-k candidates. Candidate poses are estimated with RANSAC–Kabsch \cite{fischler1981ransac, kabsch1976rotation}, and the final candidate pose is selected via minimum weighted alignment error.}
   \label{fig:pipe}
\end{figure*}

Figure \ref{fig:pipe} summarizes the Geo6DPose pipeline. In an offline onboarding stage, Geo6DPose renders RGB-D templates from the CAD models of the target objects. For each template, we extract DINOv2 patch descriptors and record the 3D coordinates of the patch centers in the rendered camera frame, producing a database of template entries $(\textbf{P},\textbf{X},T)$, where \textbf{$P$} are flattened DINO patch descriptors, $\textbf{X} \in \mathbb{R}^3$ are per-patch 3D centers, and $T$ is the camera-to-object transformation.

At inference, given a segmentation mask of a target instance, we crop and resize the input image to match the template view and extract DINOv2 patch descriptors (Sec. \ref{sec:temp}). We then compute feature similarity maps between the scene crop and the template database, and back-project scene patches to 3D using depth to form 3D–3D correspondences with template patch centers. Templates are ranked using a combined similarity and correspondence score, and the top-$k$ candidates are retained (Sec. \ref{sec:filt}). Finally, we generate pose hypotheses using RANSAC-Kabsch and select the best pose via a joint score that measures depth alignment error and mask consistency (Sec. \ref{sec:pose_hyp}).

\subsection{Template Generation}
\label{sec:temp}

Geo6DPose adopts an onboarding phase in which dense feature embeddings are precomputed for each object under a set of sampled viewpoints. To ensure consistent scale and avoid image cropping across all object orientations, the object is positioned at a fixed distance from the virtual camera. 
Given the object diameter $d$, for a target apparent size of $d_{\mathrm{px}}$ pixels, and a pinhole camera with focal lengths $f_x$ and $f_y$, the required depth $z$ is computed as
\begin{equation}
z = \frac{f \, d}{d_{\mathrm{px}}}, \quad \text{where } f = \frac{f_x + f_y}{2}.
\end{equation}
This ensures that, for every rotation, the object fits entirely within the image frame.

To cover the viewing sphere uniformly, we employ a Fibonacci sampling \cite{Marques2013SphericalFP} of the unit sphere with an angular separation of approximately $\alpha$. For each sampled direction, we further render additional poses with in-plane rotations in $\delta$ increments, achieving good coverage of $\mathrm{SO}(3)$ with $N_V$ total viewpoints.

Each render uses the dataset camera intrinsics adapted to the DINO input resolution $\mathbf{K}_R$, ensuring a $30 \times 30$ patch grid as in FoundPose~\cite{ornek2024foundpose}. For DINOv2, the patch size is $P_S = 14$px, yielding images of $I_S = 420$px.

For each render we extract RGB-D, a binary mask, and surface coordinates in the camera frame. The RGB crop is divided into non-overlapping DINO patches, yielding a per-view feature map and mask. To reduce memory and accelerate nearest-neighbor matching at test time (following prior work \cite{caraffa2024freeze, ornek2024foundpose}), Geo6DPose applies PCA per object to project patch descriptors into a compact subspace.
Let \(\mathbf{f}_{i,q}\in\mathbb{R}^D\) denote the DINO descriptor of the \(q\)-th foreground patch of view \(i\). The PCA projection is
\begin{equation}
\hat{\mathbf{f}}_{i,q} = \mathbf{W}_{\mathrm{PCA}}^\top \mathbf{f}_{i,q}, \qquad
\hat{\mathbf{f}}_{i,q} \in \mathbb{R}^{D_{\mathrm{PCA}}},
\end{equation}
where \(\mathbf{W}_{\mathrm{PCA}}\in\mathbb{R}^{D\times D_{\mathrm{PCA}}}\) is learned offline from all template descriptors of the object. Only patches whose center are intersecting the object mask are retained and flattened into a descriptor bank for view \(i\):
\begin{equation}
\mathbf{P}_i = [\,\hat{\mathbf{f}}_{i,q}\,]_{q=1}^{N_i}, \qquad
\hat{\mathbf{f}}_{i,q}\in\mathbb{R}^{D_{\mathrm{PCA}}},
\end{equation}
where \(N_i=\sum \mathbf{m}_i\) is the number of foreground patches and \(\mathbf{m}_i\) the mask generated in view \(i\) .

For each valid patch center, its 3D location is obtained via back-projection using the rendered depth $z_q$ and camera intrinsics $\mathbf{K}_R$ forming the associated set of 3D patch centers $\mathbf{X}_i$. Each template stores its local 3D patch coordinates along with the camera-to-object pose $T_i$ used to generate it, yielding the final object template database:

\begin{equation}
\mathcal{T} = \{(\mathbf{P}_i, \mathbf{X}_i, T_i)\}_{i=1}^{N_V},
\end{equation}

\vspace{0.2em}
\noindent Design choices - DINOv2-B $\rightarrow$ $I_S$ = 420px, $P_S$ = 14px. $\alpha, \delta$ = 25°, 60° $\rightarrow$ $N_V$ = 396. PCA = 256. Full ablations are provided in Tables \ref{tab:feat}-\ref{tab:templates}.

\subsection{Mutual Correspondence Filtering}
\label{sec:filt}

Given a segmented query crop at test time, Geo6DPose extracts dense DINO descriptors and back–projects the valid foreground patches into 3D, resulting in a set of $N_s$ scene patch descriptors and their 3D centers. For each template $i$ in the candidate set $\mathcal{T}$, we retrieve its PCA-compressed template descriptors and 3D centers. We compute dense pairwise cosine similarity between scene and template features using:

\begin{equation}
S_{i}(j, q) = \langle \mathbf{p}_s^j, \mathbf{p}_i^{q} \rangle, \quad S_i \in \mathbb{R}^{N_s \times N_i}.
\end{equation}

A correspondence $(j, q)$ is accepted only if it is a mutual nearest neighbor:

\begin{equation}
    \begin{aligned}
        q^*(j) &= \arg \max_q \, \mathbf{p}_s^j \cdot \mathbf{p}_i^{q}, \\
        j^*(q) &= \arg \max_j \, \mathbf{p}_s^j \cdot \mathbf{p}_i^{q}, \\
        \mathbb{M}_i(j) &= 
        \begin{cases} 
            1, & \text{if } j = j^*(q^*(j)) \\ 
            0, & \text{otherwise}
        \end{cases}
    \end{aligned}
    \label{eq:placeholder_label}
\end{equation}

where $\mathbb{M}_{i}(j)$ is the  mutual consistency mask for scene patch $j$ in template $i$. Let $C_i=\sum \, \mathbb{M}_{i}(j)$ denote the number of consistent matches and $\bar{s}_i$ their mean similarity. Each template is ranked using a trade-off between geometric support (match coverage) and descriptor confidence:

\begin{equation}
\text{score}_i = \gamma \frac{C_i}{N_s} + (1 - \gamma) \bar{s}_i,
\label{eq:score_t}
\end{equation}

where $\gamma \in[0,1]$ balances coverage vs. descriptor quality. The highest-scoring top-$k$ templates are selected and for each selected template, all valid mutual correspondences are recovered yielding 3D–3D matches between the scene and the template. These correspondence sets  are forwarded to the next stage for pose generation hypothesis.

\noindent Design choices - $\gamma =0.3$. Top-$k = 15$. Ablations provided in Table \ref{tab:topk}.

\subsection{Pose Hypothesis and Selection}
\label{sec:pose_hyp}

Given the top-$k$ retrieved template correspondences, we first map template points into the object model frame by applying the inverse of the template pose, yielding a set of scene–model correspondences.

We then estimate pose hypotheses using a fast Numba-accelerated RANSAC–Kabsch solver. Each RANSAC iteration samples three correspondences, estimates a rigid transformation $(\mathbf{R}, \mathbf{t})$ via the closed-form Kabsch SVD solution, and scores it by the inlier count~\cite{fischler1981ransac, kabsch1976rotation}. A correspondence is considered an inlier if

\begin{equation}
|\mathbf{R} c_i^T + \mathbf{t} - c_i^S|_2 < \tau,
\quad \tau = 0.05 d,
\end{equation}

where $d$ is the object diameter, and $(c_i^T, c_i^S)$ denotes a correspondence pair in model and scene coordinates, respectively.
The best hypothesis with at least six inliers is retained, forming the final pose $T = [\mathbf{R} \,|\, \mathbf{t}]$. Rather than selecting the final pose by inlier count alone, we instead score hypotheses using the proposed Weighted Alignment Error (WAE), which balances 3D alignment quality against geometric support from the observed surface:

\begin{equation}
\mathrm{WAE}(T) =
\frac{
\mathbb{E}_{i} \left[ \min\limits_{y \in S} \lVert T c_i^T - y \rVert_2 \right]
}{
\mathbb{E}_{i} \left[ \mathbf{1}\,[T x_i \in \mathcal{M}_S] \right]
},
\end{equation}

where:
\begin{itemize}
    \item $\mathbb{E}_i[\cdot]$ denotes the empirical mean
    \item $S$ is the lifted scene depth point cloud,
    \item $\mathcal{M}_S$ is the visible scene mask back-projected to 3D,
    \item $\mathbf{1}[\cdot]$ is the indicator function evaluating whether the transformed model point lies within the visible scene support.
    \item $x$ are the CAD model points in model coordinates
\end{itemize}

The numerator measures the average euclidean distance of transformed correspondences to the closest surface in the scene, while the denominator rewards hypotheses that map a larger fraction of CAD model points into the valid observed region. The final pose is selected as the pose minimizing the WAE.

\section{Experiments}
\begin{table*}[htbp]

\centering
\setlength{\tabcolsep}{2pt} 
\caption{Performance on the seven core BOP datasets. The table shows Average Recall (AR) scores per dataset, the average AR score, and the time to estimate poses of all objects in an image averaged over the datasets (in seconds). We only report methods for coarse estimation (without refinement)$^1$. Reported sub-second speed methods are in the top part of the table, while the bottom part reports training-free methods. We report best AR and average time over a set of experiments. The CNOS prior is used for all methods}
\begin{tabular}{l*{12}{c}}
\toprule
     Method & Input & GPU & train-free & LM-O & T-LESS & TUD-L & IC-BIN & ITODD & HB & YCB-V & Average & Time\\  
\midrule 
     GigaPose \cite{nguyen2024gigapose}  & RGB & V100 &  & 29.6 & 26.4 & 30.0 & 22.3 & 17.5& 34.1 & 27.8 & 26.8 & 0.38s\\
     PicoPose \cite{liu2025picopose} & RGB & 3090 & & 46.3 & 39.7 & 53.6 & 36.4 & 31.0 & 66.5 & 58.7 & 47.5 & 0.51s\\
\rowcolor{yellow!20} Geo6DPose (ours) & RGB-D & A4000 & \checkmark & 55.2 & 47.8 & 66.3 & 43.0 & 40.2 & 68.9 & 59.8 & 53.7 & 0.92s\\
     FoundPose \cite{ornek2024foundpose} & RGB & P100 & \checkmark & 39.7 & 33.8 & 46.9 & 23.9 & 20.4 & 50.8 & 45.2 & 37.3 & 1.69s\\
     FreeZeV2$^1$ \cite{caraffa2025freezev2}  & RGB-D & A40 & \checkmark & 69.4 & 56.3 & 93.5 & 54.5 & 60 & 76.9 & 87.4 & 64.0 & 1.5s\\
     ZS6D \cite{ausserlechner2024zs6d}  & RGB & - & \checkmark & 29.8 & 21.0 & - & - & - & - & 32.4 & - & -\\
     DZOP  \cite{von2024dzop} & RGB & - & \checkmark & 32.8 & 26.7 & - & - & - & - & 45.2 & - & -\\  
\bottomrule
    
\end{tabular}
\label{tab:SOTA}
\end{table*}

We evaluate Geo6DPose following the \textit{BOP Challenge 6D localization of unseen objects} evaluation protocol . Each method is tasked with estimating the 6D pose of object instances in RGB-D images with known camera intrinsics, CAD models, number of instances per object-ID, and segmentation masks provided by an external tool \cite{hodan2018bop}.

\subsection{Evaluation Protocol, Datasets and Experimental Setup}
Following BOP conventions, we report pose accuracy using the three canonical error functions: 

\begin{enumerate}
    \item Visible Surface Discrepancy (VSD): measures the alignment of rendered object masks against the visible parts in the observed depth.
    \item Maximum Symmetry-Aware Surface Distance (MSSD): evaluates the maximal distance between the estimated and ground-truth object surfaces, accounting for symmetries.
    \item Maximum Symmetry-Aware Projection Distance (MSPD): measures the maximal pixel projection error of object points under the estimated pose, considering symmetries.
\end{enumerate}

An estimated pose is considered correct if its error falls below a predefined threshold for each metric. We aggregate results using the Average Recall (AR), computed as the mean of the recall values across the three error functions \cite{hodan2018bop}:
$$AR=\frac{1}{3}(AR_{VSD} + AR_{MSSD} + AR_{MSPD})$$

Experiments are conducted on the 7 core BOP datasets: LM-O \cite{Brachmann_lmo}, T-LESS \cite{hodan2017tlessrgbddataset6d}, ITODD \cite{drost2017itodd}, HB \cite{kaskman2019homebreweddb}, YCB-V \cite{xiang2018posecnn}, IC-BIN \cite{kleeberger2019icbin}, and TUD-L \cite{hodan2018bop}, which provide a variety of object textures, shapes, and cluttered scenes. These datasets are widely used for benchmarking 6D pose estimation methods.

All experiments were conducted on a workstation equipped with an AMD Ryzen 7-7700X CPU and a NVIDIA RTX A4000 GPU (16 GB VRAM). Each experiment runs on a single GPU with models in evaluation mode and no mixed precision.
\subsection{Main Results}
\label{sec:main}

Among training-free methods, Geo6DPose achieves average recall (AR) competitive with the current state-of-the-art (SOTA) FreeZeV2, which relies on a large backbone combining DINOv2-G with the proprietary GeDi transformer. Although FreeZeV2\footnote{FreeZev2 scores include ICP refinement, since results without refinement were not provided.} reports strong accuracy, the complexity of its architecture limits its suitability for real-time applications when compared to sub-second methods such as GigaPose, PicoPose, and our approach, Geo6DPose. We note that FreeZeV2’s public AR numbers include ICP refinement (as unrefined results are not available), whereas Geo6DPose and all other methods in Table~\ref{tab:SOTA} report results without ICP or any pose refinement. Despite this, FreeZev2 achieves 19\% better AR while running 56\% slower.

Compared to FoundPose, which also operates without training and served as inspiration for our approach, Geo6DPose delivers higher AR despite using smaller DINO backbones. We hypothesize that this improvement stems from leveraging depth information to better validate pose hypotheses, enabled by our RANSAC + WAE scoring mechanism, which provides more reliable pose filtering than FoundPose’s PnP-RANSAC inlier-count criterion.

Finally, real-time capable baselines GigaPose and PicoPose, achieves substantial AR detriments performing at 50\% and 12\% lower against Geo6DPose, respectively, while Geo6DPose runtime also exceeds 1 FPS, making it both more accurate and suitable for interactive robotics perception.

For Table~\ref{tab:SOTA}, we follow the CNOS-recommended evaluation protocol~\cite{nguyen2023cnos}, filtering detections by valid object IDs, expected instance counts, and bounding box overlap with the ground-truth amodal boxes using an IoU threshold to prevent pose estimation on background regions. This procedure is functionally equivalent to the evaluation code released by FoundPose~\cite{ornek2024foundpose}. Table~\ref{tab:prior} emulates a no-target setting, where ground-truth annotations are not used to verify whether a CNOS detection overlaps with a target object. Instead, detections are filtered only by CNOS confidence, reflecting a realistic deployment of the full CNOS+Geo6DPose pipeline. For reference, we also report results using BOP target lists and ground-truth masks.

\begin{table}[h]
\centering
\setlength{\tabcolsep}{2pt} 
\caption{Average Recall (AR) of Geo6Dpose on the seven core BOP datasets using different CNOS prior protocol and ground truth (GT) masks. CNOS-tgt refers to the use of number of instances per object as a filter. CNOS-conf uses only CNOS confidence $>$ 0.5 as a filter.}
\begin{tabular}{l*{10}{c}}
\toprule
     Prior Protocol & \rotatebox{90}{LM-O} & \rotatebox{90}{T-LESS} & \rotatebox{90}{TUD-L} & \rotatebox{90}{IC-BIN} & \rotatebox{90}{ITODD} & \rotatebox{90}{HB} & \rotatebox{90}{YCB-V} & \rotatebox{90}{Avg.}\\  
\midrule 
     CNOS-conf  &  47.2 & 45.1 & 67.5 & 43.0 & 41.7 & 58.3 & 59.6 & 51.8 \\
     CNOS-tgt  &  55.2 & 47.8 & 66.3 & 43.0 & 40.2 & 68.9 & 59.8 & 53.7\\
     GT &  52.7 & 78.4 & 76.2 & 49.2 & - & - & 58.3 & - \\
     FoundPose-GT  &  45.6 & 53.1 & 57.1 & 30.6 & - & - & 50.9 & - \\

\bottomrule
\end{tabular}
\label{tab:prior}
\end{table}

We observe a clear degradation in performance in HB and LM-O datasets when instance-level priors are removed. Although we implemented a safeguard to discard scene point clouds with fewer than 20 points —an event that was empirically triggered frequently— the runtime on datasets such as TUD-L and IC-BIN increased by nearly a factor of five. This suggests that a substantial amount of CNOS noise or irrelevant background was still being processed during pose estimation. Increasing the confidence threshold and introducing additional filters —such as rejecting masks whose projected size exceeds the expected object diameter at a given depth— could further suppress background regions and reduce the overhead from spurious pose hypotheses.

\subsection{Ablation Experiments}
\label{sec:abla}
\textit{Runtime values reported in this section exclude CNOS overhead to allow controlled, method-focused comparisons across design choices. End-to-end timings including CNOS are reported in Fig.~\ref{fig:abla}.}\\

The final pipeline configuration was determined through a systematic three-stage ablation study. Unless otherwise noted, the baseline configuration uses: \textit{DINOv3-S features, no PCA compression, final backbone layer (12), 618 pre-rendered templates, and top-15 correspondence retrieval}. We then analyze the impact of (i) feature extraction design, (ii) template sampling density, and (iii) correspondence filtering, selecting configurations that best balance pose accuracy, runtime efficiency, and memory footprint.

\subsubsection{Feature Extractors}
We first evaluate the influence of backbone architecture, PCA compression, and transformer layer selection on the accuracy–runtime trade-off. Across all model scales, DINOv2 consistently outperforms DINOv3, indicating stronger correspondence-preserving representations for geometric matching. PCA compression reliably reduces runtime, though larger backbones require higher retained dimensionality to avoid degrading accuracy.

\begin{table}[ht]

\centering
\setlength{\tabcolsep}{2pt} 
\caption{Ablation of feature extraction design. We evaluate PCA dimensionality (top) and DINO layer selection (bottom) across multiple DINO model variants to identify the best trade-off between average runtime per image $t$ (s) and Average Recall $AR$ (\%). All experiments are conducted on the LM-O dataset \cite{Brachmann_lmo} using CNOS segmentation priors \cite{nguyen2023cnos}.}
\begin{tabular}{l*{10}{c}}

\toprule
\multicolumn{11}{c}{PCA Sensitivity Analysis}\\
\midrule
     Model & \multicolumn{3}{c}{DINOv3s} & \multicolumn{3}{c}{DINOv3b} & \multicolumn{2}{c}{DINOv2s} & \multicolumn{2}{c}{DINOv2b}\\
     PCA & 64 & 128 & - & 128 & 256 & - & 128 & - & 256 &- \\ 
     $AR$ & 41.1 & 46.2 & 48.9 & 49.3 & 51.1 & 52.5 & 44.7 & 48.7 & 53.0 & 53.6 \\ 
     $t$ & 0.93 & 0.95 & 1.05 & 1.03 & 1.10 & 1.33 & 0.96 & 1.09 & 1.15 & 1.41 \\ 
\midrule
\multicolumn{11}{c}{Layer Output Sensitivity Analysis}\\
\midrule
     Model & \multicolumn{3}{c}{DINOv3s} & \multicolumn{3}{c}{DINOv3b} & \multicolumn{2}{c}{DINOv2s} & \multicolumn{2}{c}{DINOv2b}\\
     PCA & \multicolumn{3}{c}{128} & \multicolumn{3}{c}{256} & \multicolumn{2}{c}{128} & 128 & \textbf{256}\\
     Layer & 12 & 11 & 9 & 12 & 11 & 9 & 12 & 9 & \multicolumn{2}{c}{\textbf{9}} \\ 
     $AR$ & 46.2 & 49.2 & 49.6 & 51.1 & 47.6 & 53.2 & 44.7 & 51.7 & 52.6 & \textbf{55.7} \\ 
     $t$ & 0.95 & 0.96 & 0.98 & 1.10 & 1.08 & 1.12 & 0.96 & 1.00 & 1.11 & \textbf{1.18} \\ 

\bottomrule
    
\end{tabular}
\label{tab:feat}
\end{table}

Consistent with findings in FoundPose \cite{ornek2024foundpose}, mid-level transformer layers (9–11) outperform the final layer, suggesting that intermediate features better preserve local spatial structure, whereas last-layer embeddings become overly abstract for correspondence search. In our setting, layer 9 delivers, on average, an +8\% AR gain compared to the final layer of the same backbone.

While larger models increase latency, DINOv2-B (layer 9, PCA=256) emerges as the strongest high-accuracy variant, improving AR by 7\% over DINOv2-S (layer 9, PCA=128) with a runtime increase of only 0.18s per image. Conversely, DINOv2-S (layer 9, PCA=128) retains the most favorable speed–accuracy trade-off among the fastest configurations.

Overall, these results identify DINOv2-S (PCA=128, layer 9) and DINOv2-B (PCA=256, layer 9) as the most compelling feature representations, forming strong Pareto-optimal operating points for efficient and accurate pose estimation.

\subsubsection{Pipeline}
We next study how template sampling density affects recall, runtime, and storage, controlling $SO(3)$ coverage via viewpoint spacing $\alpha$ on the Fibonacci sphere and in-plane rotation step $\delta$ (see Table \ref{tab:templates}. Denser sampling increases recall but at substantial computational and memory cost. Extremely dense configurations (e.g., $\alpha$=15°, $\delta$=45°) offer only marginal AR gains over moderate sampling, yet incur a $>$2.3× increase in runtime and over 5× larger on-disk descriptors, making them unsuitable for practical deployment. In contrast, overly sparse sampling ($\alpha\ge45$°) sharply reduces inference time but leads to a pronounced decline in AR, indicating insufficient view coverage to ensure stable pose retrieval.
\begin{table}[ht]
  \caption{Onboarding ablation study. We analyze the effect of template ($N$) density on performance, controlled by camera sampling in $SO(3)$. Views are placed on a Fibonacci sphere with angular spacing $\alpha$, and each view is augmented with in-plane rotations at step $\delta$. Experiments are conducted on LM-O using CNOS masks.}
  
  \centering
  \setlength{\tabcolsep}{4pt} 
  \begin{tabular}{l*{5}{c}}
    \toprule
    Model-PCA & $\alpha$ & $\delta$ & $N_V$ & $AR$ & $t$ \\
    \midrule
    DINOv2s-128 & 15° & 45° & 1464 & 53.1 & 1.80\\ 
    DINOv2s-128 & 30° & 60° & 270 & 49.1 & 0.73 \\
    DINOv2s-128 & 45° & 60° & 120 & 43.6 & 0.53 \\
    \textbf{DINOv2b-256} & \textbf{25°} & \textbf{60°} & \textbf{396} & \textbf{55.2} & \textbf{0.95}\\ 
    DINOv2b-256 & 30° & 45° & 360 & 53.7 & 0.9\\ 
    DINOv2b-128 & 30° & 45° & 360 & 50.7 & 0.85\\ 
    
    \bottomrule
  \end{tabular}
  \label{tab:templates}
\end{table}

The configuration $\alpha$=25°, $\delta$=60° (396 templates) achieves the best overall trade-off, improving AR by +1.5 points for only +0.05 s additional runtime. Further gains show diminishing returns: increasing to $\alpha$=20°, $\delta$=60° (618 templates, Table~\ref{tab:feat}) yields only half additional AR point while increasing inference time by 25\% and doubling the descriptor storage footprint.

Inspired by FoundPose’s bag-of-words pre-retrieval strategy \cite{ornek2024foundpose}, Table \ref{tab:topk} investigates a preliminary filtering stage using dense DINO similarity maps to prune templates prior to sparse correspondence matching, with the goal of discarding unlikely poses before the more expensive top-$k$ geometric stage. However, another set of dense DINO features must be stored per template, doubling onboarding compute and storage. Furthermore, this design requires two backbone forward passes at inference (dense filtering $\rightarrow$ sparse matching), which adds latency even when the candidate set is reduced.

In practice, dense pre-filtering (Dense-$k$ + Sparse-$k$) provides no recall gains and consistently introduces a 30–45\% runtime increase over sparse-only retrieval, which remains efficient with moderate $k$ values. Given the higher memory footprint, increased onboarding cost, and inferior accuracy–time trade-off, we discard dense pre-filtering and adopt a single-stage sparse top-$k$ strategy, which proves both faster and more accurate in our pipeline.

\begin{table}[t]
  \caption{Top-k filtering ablation study. We evaluate the impact of top-k similarity selection on performance and study an additional template pre-filtering stage for efficiency gains. Results are reported on LM-O using CNOS masks.}
  
  \centering
  \setlength{\tabcolsep}{4pt} 
  \begin{tabular}{l*{4}{c}}
    \toprule
    Model-PCA-$N_V$ & Dense-k & Sparse-k & $AR$ & $t$\\
    \midrule
    DINOv2s-128-270 & - & 20 & 49.6 & 0.75  \\
    DINOv2s-128-270 & 30 & 15 & 48.6 & 1.10  \\
    DINOv2s-128-270 & 20 & 10 & 47.8 & 0.98  \\
    DINOv2s-128-270 & - & 1 & 31.0 & 0.47  \\
    \textbf{DINOv2b-256-396} & - & \textbf{10} & \textbf{53.6} & \textbf{0.86}  \\
    DINOv2b-256-396 & - & 5 & 52.7 & 0.79  \\
    DINOv2b-256-396 & - & 1 & 41.0 & 0.73  \\
    DINOv2b-128-360 & - & 10 & 49.5 & 0.78  \\
    
    \bottomrule
  \end{tabular}
  \label{tab:topk}
\end{table}

When limiting retrieval to sparse-only matching (no dense pre-filtering), the choice of $k$ exposes a predictable accuracy–speed trade-off, where $k=1$ marks the maximum inference speed achievable by each configuration. Moderate pruning can reduce runtime with small AR degradation, but gains diminish quickly: Taking DINOv2b-256-396, reducing $k$ from 15 to 10 lowers AR by 1.6 points while saving only 110 ms, indicating an unfavorable trade-off.

\subsubsection{Candidate Pose Selection}

The final ablation available in Table \ref{tab:final} assesses the benefit of our Weighted Alignment Error (WAE) pose ranking against the common RANSAC heuristic of selecting the hypothesis with the highest inlier count, a criterion adopted in prior 6D estimators such as FoundPose \cite{ornek2024foundpose}. While max-inlier selection ignores geometric support and reprojection quality, WAE explicitly scores both, favoring poses that explain more 3D structure and align more consistently in image space, albeit requiring RGB-D input. Although WAE introduces an additional ~50 ms per frame, it yields a +2 AR improvement, validating the value of richer geometric pose verification despite the small runtime overhead.

\begin{table}[ht]
  \caption{Comparison of pose selection criteria reporting Average Recall $AR$ (\%) and inference time $t$ (s) on LM-O with CNOS masks.}
  
  \centering
  \setlength{\tabcolsep}{4pt} 
  \begin{tabular}{l*{4}{c}}
    \toprule
    Model-PCA-$N_V$ & $k$ & Pose Selector &  $AR$ & $t$\\
    \midrule
    DINOv2b-256-396 & 15 & Max. Inliers & 53.2 & 0.87\\
    \textbf{DINOv2b-256-39}6 & \textbf{15} & \textbf{WAE} & \textbf{55.2} & \textbf{0.95}\\
    \bottomrule
  \end{tabular}
  \label{tab:final}
\end{table}

\subsubsection{Final Remarks}

\begin{figure}[t]
  \centering
   \includegraphics[width=\linewidth]{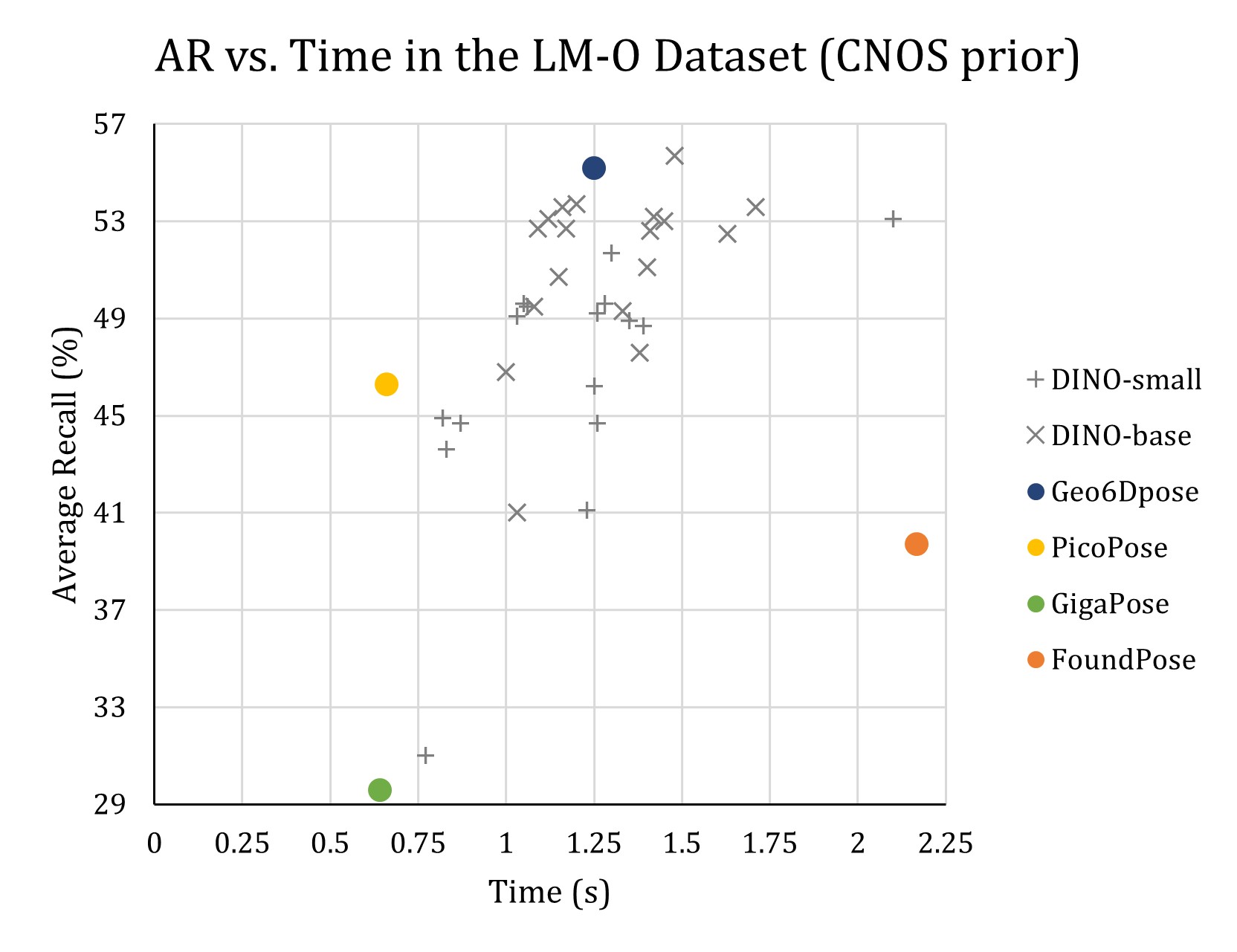}

   \caption{Geo6Dpose ablation studies from Tables \ref{tab:feat}-\ref{tab:topk} represented in a scatter AR-time pareto plot compared against SOTA subsecond methods. Times and AR reported for LM-O using CNOS masks.}
   \label{fig:abla}
\end{figure}

The AR–time distribution in Fig.~\ref{fig:abla} summarizes the practical trade-offs explored across our design space. Within this landscape, Geo6DPose (55.2 AR, 0.95s) lies on the Pareto frontier of our ablation sweep. For context, FoundPose~\cite{ornek2024foundpose}, PicoPose~\cite{liu2025picopose}, and GigaPose~\cite{nguyen2024gigapose} are RGB-only methods, with the latter two relying on training, while both FoundPose and Geo6DPose are training-free, the latter additionally operating on RGB-D input. LM-O, offering strong RGB texture and noisy depth, remains a fair benchmark for both modalities due to its realistic sensing conditions. Within the RGB-D category, two recent BOP leaderboard entries (without published descriptions) report both higher AR and faster inference than Geo6DPose, albeit requiring training, confirming that further gains remain attainable.

Even so, Fig.~\ref{fig:abla} highlights two key takeaways: (1) Geo6DPose markedly outperforms FoundPose (39.7 AR, 2.16s) and (2) against trained RGB baselines, it surpasses PicoPose (46.3 AR, 0.66s) and significantly outperforms GigaPose (29.6 AR, 0.64s), while operating within 300 ms of their runtimes despite being benchmarked on comparatively lower-end hardware (RTX A4000, 16 GB) versus RTX 3090~\cite{liu2025picopose} and Tesla V100~\cite{nguyen2024gigapose}. Overall, the results indicate that template-based, training-free RGB-D methods can attain competitive operating points in the AR–FPS space, without requiring learned models, network access, or large compute budgets.

Finally, Geo6DPose currently operates without a learned refinement stage, as most non-geometric refiners introduce latency incompatible with robotic requirements. Future work will explore efficient real-time refinement; further optimization of the Python pipeline to remove residual overheads; and deployment on articulated robots, including on-the-fly 6D scene reconstruction using an eye-in-hand depth sensor.

\label{sec:exps}
\section{Conclusion}
\label{sec:conc}
We presented Geo6DPose, a fast, fully local, and training-free framework for zero-shot 6D object pose estimation. By combining foundation-model visual features with geometric filtering and 3D–3D correspondence reasoning, Geo6DPose achieves competitive accuracy to state-of-the-art trained and train-free models while maintaining real-time capabilities on a single commodity GPU. Our experiments demonstrate that strong geometric priors can effectively compensate for the absence of learned refinement, enabling practical on-device inference without compromising robustness. The proposed weighted geometric alignment metric further improves stability under clutter and occlusion. Overall, Geo6DPose advances the efficiency–accuracy trade-off for zero-shot 6D perception and offers a promising direction for scalable, training-free vision in robotic applications.
{
    \small
    \bibliographystyle{ieeenat_fullname}
    \bibliography{main}
}

\clearpage
\setcounter{page}{1}
\maketitlesupplementary

%
%
%

\section{Qualitative Results}
\begin{figure*}[ht]
  \centering
   \includegraphics[width=0.9\linewidth]{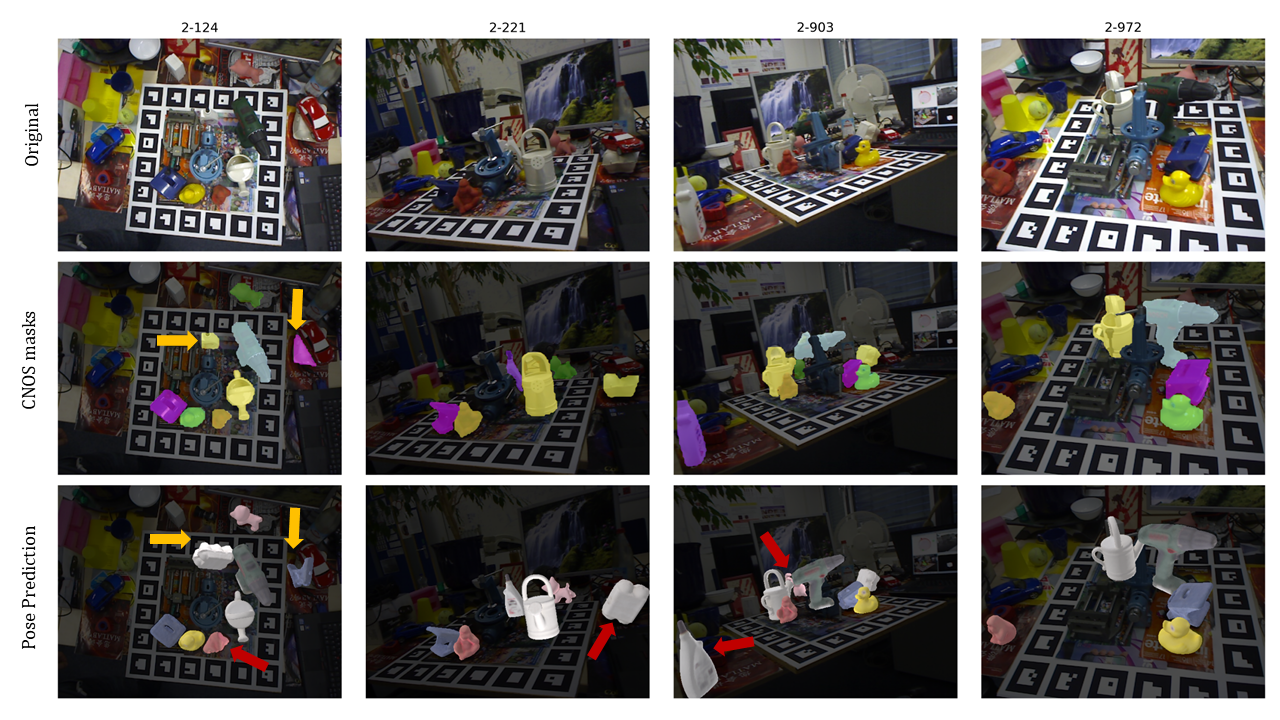}

\caption{
    Qualitative results on the \textbf{LM-O dataset} \cite{Brachmann_lmo}. Scene ID and Image ID are provided for each instance.
    \textbf{Top:} original images. 
    \textbf{Middle}: CNOS masks filtered by target objects. 
    \textbf{Bottom:} final pose predictions. 
    \textcolor{yellow}{\Large$\bm{\Rightarrow}$}~Yellow arrows indicate CNOS class-prediction errors and how these errors propagate into the final pose. \textcolor{red}{\Large$\bm{\Rightarrow}$}~Red arrows highlight pose-estimation errors produced by Geo6DPose.}
   \label{fig:lmo}
\end{figure*}

In this section, we present qualitative examples from all BOP benchmark datasets used to evaluate Geo6DPose. For each dataset, we show four samples, including the CNOS predictions filtered by the target classes and the resulting pose estimations. Yellow arrows highlight CNOS class-prediction errors, which unavoidably propagate to Geo6DPose. Red arrows indicate different types of pose errors or misalignments, pointing to failure modes that can guide future improvements. Below, we provide dataset-specific observations:

\noindent \textbf{LM-O Dataset} See Figure \ref{fig:lmo}. This dataset is where we conducted the most extensive set of experiments. In images \textit{2-124}, \textit{2-221}, and \textit{2-903}, the misalignments of the \textit{ape}, \textit{egg tray}, and \textit{cat} objects primarily stem from challenging viewpoints that produce ambiguous contours or self-occlusions.

In contrast, the misalignment of the glue bottle in image \textit{2-903} arises from a fallback case in Geo6DPose: the depth signal within the segmented region is too sparse for reliable correspondence matching. When this happens, the method defaults to placing the best-scoring template at the median depth and centered within the mask. RGB-based refiners \cite{labbé2022megapose, moon2024genflow} would likely be able to recover this pose.

\begin{figure*}[ht]
  \centering
   \includegraphics[width=0.9\linewidth]{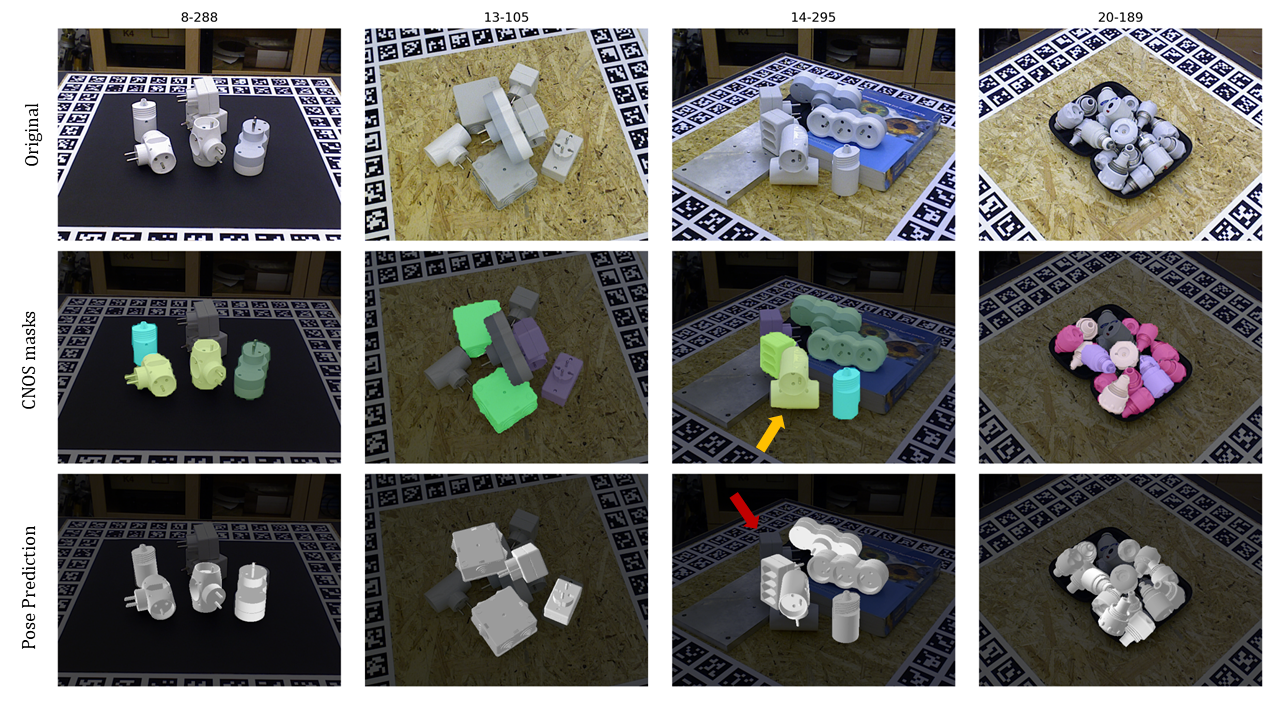}

\caption{
    Qualitative results on the \textbf{T-LESS dataset} \cite{hodan2017tlessrgbddataset6d}. Scene ID and Image ID are provided for each instance.
    \textbf{Top:} original images. 
    \textbf{Middle}: CNOS masks filtered by target objects. 
    \textbf{Bottom:} final pose predictions. 
    \textcolor{yellow}{\Large$\bm{\Rightarrow}$}~Yellow arrows indicate CNOS class-prediction errors and how these errors propagate into the final pose. \textcolor{red}{\Large$\bm{\Rightarrow}$}~Red arrows highlight pose-estimation errors produced by Geo6DPose.}
   \label{fig:tless}
\end{figure*}

\noindent \textbf{T-LESS Dataset} See Figure \ref{fig:tless}. In this dataset, most pose errors originate from the mask-prediction stage. This is supported by the large performance gap between using ground-truth masks for the BOP targets (78.4\% AR) versus using CNOS masks (47.8\% AR) (See Table \ref{tab:prior}). Image \textit{14-295} illustrates a complete mask omission, which is linked to an unexpected code exit triggered during the WAE computation. The underlying cause of this failure is still under investigation.

\begin{figure*}[ht]
  \centering
   \includegraphics[width=0.9\linewidth]{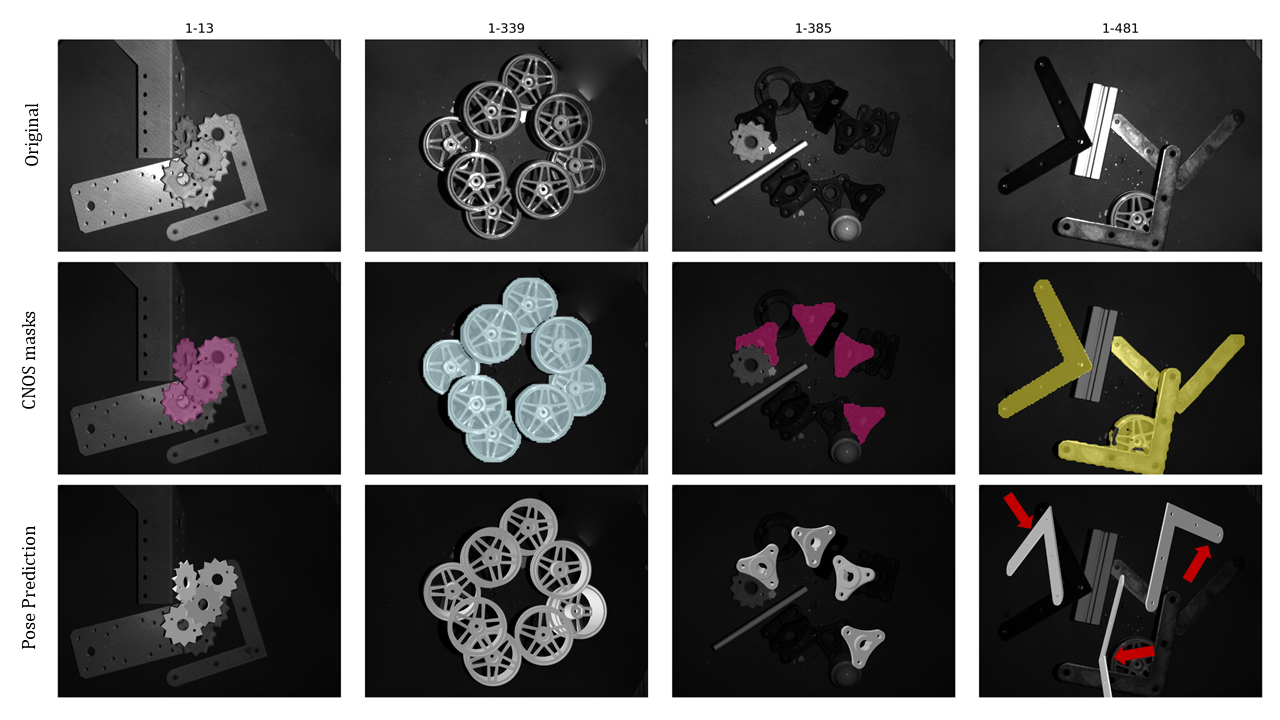}

\caption{
    Qualitative results on the \textbf{ITODD dataset} \cite{drost2017itodd}. Scene ID and Image ID are provided for each instance.
    \textbf{Top:} original images. 
    \textbf{Middle}: CNOS masks filtered by target objects. 
    \textbf{Bottom:} final pose predictions. 
    \textcolor{yellow}{\Large$\bm{\Rightarrow}$}~Yellow arrows indicate CNOS class-prediction errors and how these errors propagate into the final pose. \textcolor{red}{\Large$\bm{\Rightarrow}$}~Red arrows highlight pose-estimation errors produced by Geo6DPose.}
   \label{fig:itodd}
\end{figure*}

\noindent \textbf{ITODD Dataset} See Figure \ref{fig:itodd}.The ITODD dataset presents particularly challenging scenarios for RGB‐based algorithms trained on rich texture, an inherent limitation for methods such as FoundPose, PicoPose, and GigaPose. In contrast, performance increases substantially when depth information is incorporated, as demonstrated by both Geo6DPose and FreeZe. 
However, image \textit{1-481} exposes an unusual failure mode of Geo6DPose on planar, texture-poor sheet-metal parts. Due to the lack of reliable correspondences, Geo6DPose falls back to the best-scoring template, mirroring the behavior observed in LM-O image \textit{2-903}. 
Such poses are difficult for any refiner to recover and highlight a fundamental limitation of Geo6DPose—and, more broadly, of any method relying heavily on DINO-based visual features.

\begin{figure*}[ht]
  \centering
   \includegraphics[width=0.9\linewidth]{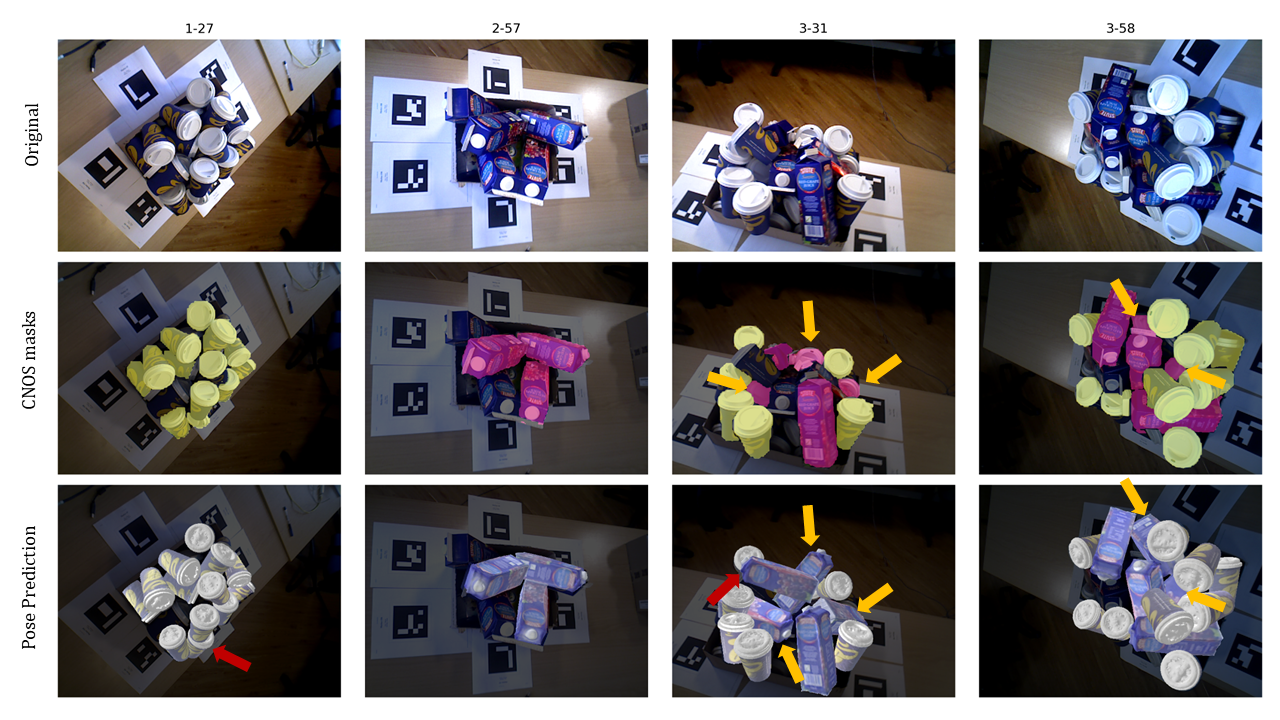}

\caption{
    Qualitative results on the \textbf{IC-BIN dataset} \cite{kleeberger2019icbin}. Scene ID and Image ID are provided for each instance.
    \textbf{Top:} original images. 
    \textbf{Middle}: CNOS masks filtered by target objects. 
    \textbf{Bottom:} final pose predictions. 
    \textcolor{yellow}{\Large$\bm{\Rightarrow}$}~Yellow arrows indicate CNOS class-prediction errors and how these errors propagate into the final pose. \textcolor{red}{\Large$\bm{\Rightarrow}$}~Red arrows highlight pose-estimation errors produced by Geo6DPose.}
   \label{fig:icbin}
\end{figure*}

\noindent \textbf{IC-BIN Dataset} See Figure \ref{fig:icbin}. As in T-LESS, most errors in this dataset originate from the mask-prediction stage. 
However, unlike T-LESS, using ground-truth annotations yields only a marginal improvement in AR (See Table \ref{tab:prior}). 
This is largely because the ground-truth masks contain many heavily occluded objects, which are difficult for CNOS to detect and equally challenging for any method to estimate poses from.

\begin{figure*}[ht]
  \centering
   \includegraphics[width=0.9\linewidth]{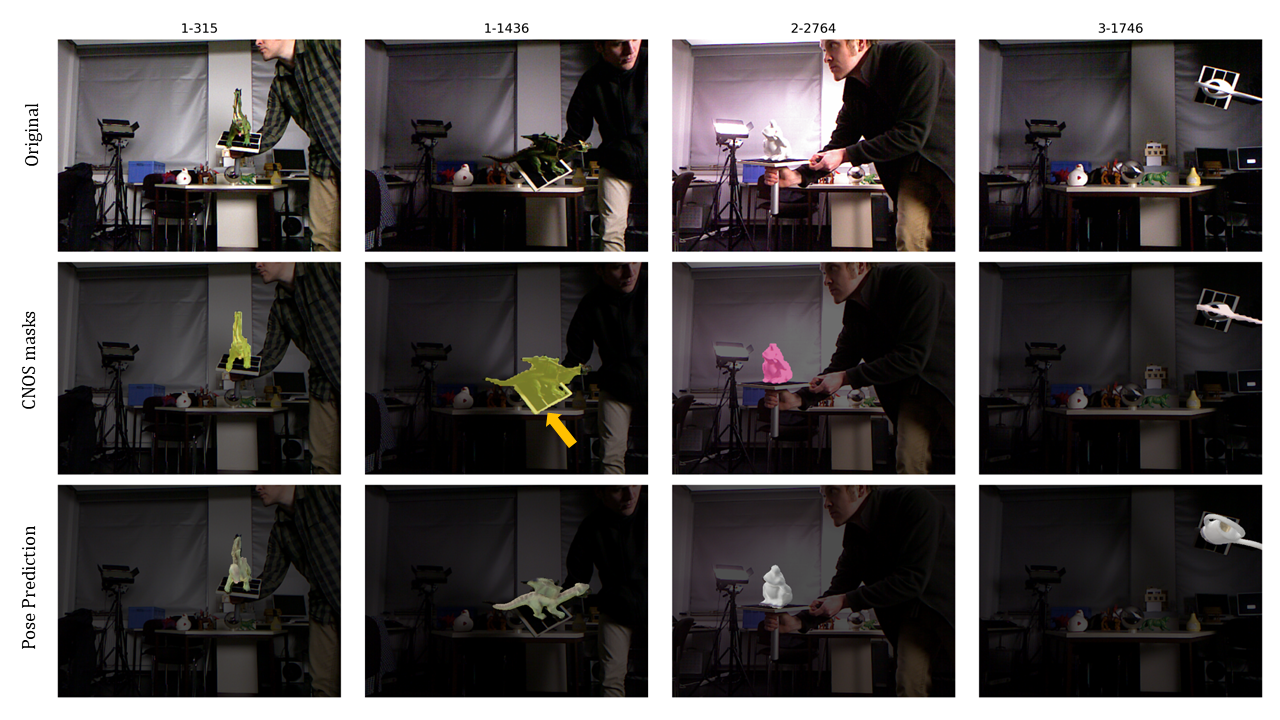}

\caption{
    Qualitative results on the \textbf{TUD-L dataset} \cite{hodan2018bop}. Scene ID and Image ID are provided for each instance.
    \textbf{Top:} original images. 
    \textbf{Middle}: CNOS masks filtered by target objects. 
    \textbf{Bottom:} final pose predictions. 
    \textcolor{yellow}{\Large$\bm{\Rightarrow}$}~Yellow arrows indicate CNOS class-prediction errors and how these errors propagate into the final pose. \textcolor{red}{\Large$\bm{\Rightarrow}$}~Red arrows highlight pose-estimation errors produced by Geo6DPose.}
   \label{fig:tudl}
\end{figure*}

\noindent \textbf{TUD-L Dataset} See Figure \ref{fig:tudl}. The main challenge in this dataset is the presence of extreme illumination conditions, which significantly affects RGB-only methods—even when depth is incorporated during pose retrieval, as in Geo6DPose. 
This explains why methods that fully exploit depth information, such as FreeZe, achieve near-perfect results, while RGB-based approaches fall behind. 
Although Geo6DPose leverages depth to correct DINO-based correspondences, certain objects still exhibit mild misalignments, as seen in image \textit{3-1746}. 
We expect that geometric refiners such as ICP could substantially reduce these residual errors and further improve pose accuracy.

\begin{figure*}[ht]
  \centering
   \includegraphics[width=0.9\linewidth]{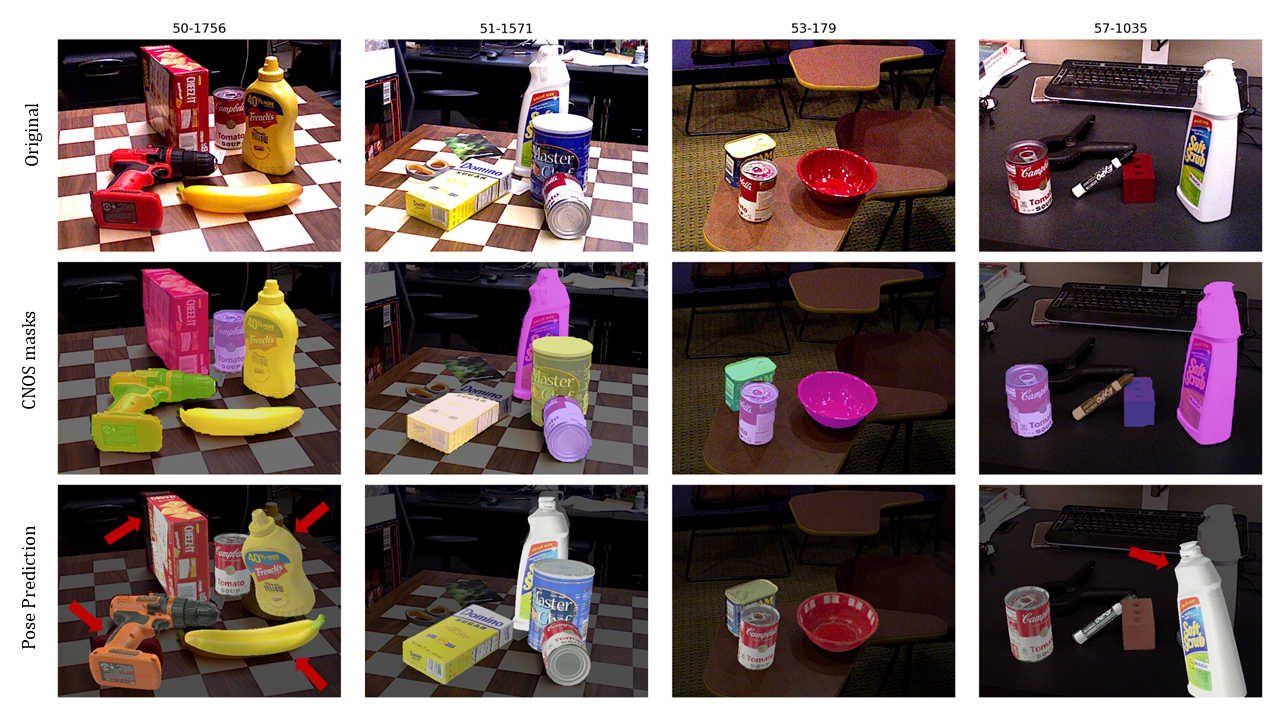}

\caption{
    Qualitative results on the \textbf{YCB-V dataset} \cite{xiang2018posecnn}. Scene ID and Image ID are provided for each instance.
    \textbf{Top:} original images. 
    \textbf{Middle}: CNOS masks filtered by target objects. 
    \textbf{Bottom:} final pose predictions. 
    \textcolor{yellow}{\Large$\bm{\Rightarrow}$}~Yellow arrows indicate CNOS class-prediction errors and how these errors propagate into the final pose. \textcolor{red}{\Large$\bm{\Rightarrow}$}~Red arrows highlight pose-estimation errors produced by Geo6DPose.}
   \label{fig:ycbv}
\end{figure*}

\noindent \textbf{YCB-V Dataset} See Figure \ref{fig:ycbv}. This dataset contains the richest textures in the BOP benchmark, which boosts RGB-based performance substantially—improving AR by more than 20 points compared to textureless (T-LESS) and grayscale (ITODD) datasets for all RGB methods. 
However, we observe the same fallback issue in image \textit{57-1035} as in LM-O image \textit{2-903}, here affecting the \textit{soft scrub bottle}. 
A more intriguing misalignment appears in image \textit{50-1756}, where both the \textit{drill} and the \textit{cereal box} exhibit correct rotations but a consistent depth offset. 
The cause of this particular failure mode remains unclear and is a subject for further investigation.

\begin{figure*}[ht]
  \centering
   \includegraphics[width=0.9\linewidth]{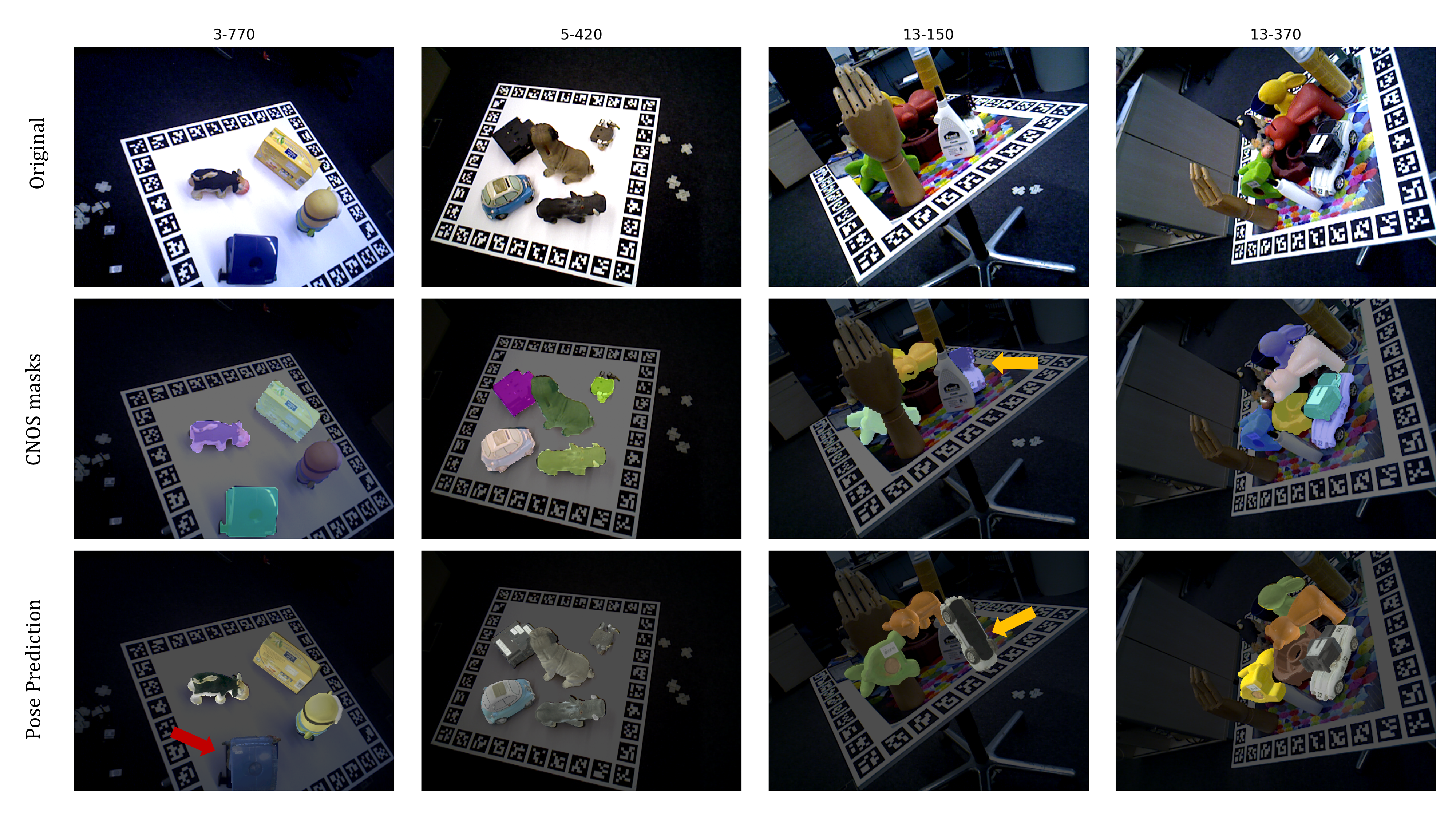}

\caption{
    Qualitative results on the \textbf{HB dataset} \cite{kaskman2019homebreweddb}. Scene ID and Image ID are provided for each instance.
    \textbf{Top:} original images. 
    \textbf{Middle}: CNOS masks filtered by target objects. 
    \textbf{Bottom:} final pose predictions. 
    \textcolor{yellow}{\Large$\bm{\Rightarrow}$}~Yellow arrows indicate CNOS class-prediction errors and how these errors propagate into the final pose. \textcolor{red}{\Large$\bm{\Rightarrow}$}~Red arrows highlight pose-estimation errors produced by Geo6DPose.}
   \label{fig:hb}
\end{figure*}

\noindent \textbf{HB Dataset} See Figure \ref{fig:hb}. The results on the HB dataset are consistent with those observed in the previous experiments. In image \textit{3-770}, the pose prediction for the \textit{perforator} is incorrect, showing a 180-degree rotation in the image plane. In image \textit{13-150}, although CNOS correctly identifies the object, the predicted \textit{car} mask is partially contaminated by a neighboring object; as a result, Geo6DPose is unable to recover the correct pose.

\end{document}